\documentclass[10pt,twocolumn,letterpaper]{article}

\usepackage{cvpr}              

\usepackage{graphicx}
\usepackage{amsmath}
\usepackage{amssymb}
\usepackage{booktabs}
\usepackage{booktabs}
\usepackage{hhline}
\usepackage[table,xcdraw]{xcolor}
\usepackage{float}
\usepackage{xcolor}
\definecolor{KleinBlue}{rgb}{0.0, 0.129, 0.7}
\definecolor{KleinRed}{rgb}{0.8, 0.129, 0.0}

\usepackage[pagebackref,breaklinks,colorlinks,citecolor=KleinBlue,linkcolor=KleinRed]{hyperref}

\usepackage[capitalize]{cleveref}
\crefname{section}{Sec.}{Secs.}
\Crefname{section}{Section}{Sections}
\Crefname{table}{Table}{Tables}
\crefname{table}{Tab.}{Tabs.}

\def\cvprPaperID{1197} 
\def\confName{CVPR}
\def\confYear{2023}
\usepackage{ulem}
\begin{document}

\title{Look, Listen, and Attack: Backdoor Attacks Against Video Action Recognition}

\author{Hasan Abed Al Kader Hammoud\textsuperscript{1} \quad
Shuming Liu\textsuperscript{1}\quad\\
Mohammed Alkhrashi\textsuperscript{2}\quad
Fahad AlBalawi\textsuperscript{2}\quad
Bernard Ghanem\textsuperscript{1}\quad\\
$^1$ King Abdullah University of Science and Technology (KAUST), Thuwal, Saudi Arabia\\
$^2$ Saudi Data and Artificial Intelligence Authority (SDAIA), Riyadh, Saudi Arabia\\
{\tt\small  \{hasanabedalkader.hammoud,shuming.liu,bernard.ghanem\} @kaust.edu.sa}\\
 {\tt\small \{mkhrashi,falbalawi\} @sdaia.gov.sa}
}

\maketitle

\begin{abstract}

Deep neural networks (DNNs) are vulnerable to a class of attacks called ``backdoor attacks", which create an association between a backdoor trigger and a target label the attacker is interested in exploiting. A backdoored DNN performs well on clean test images, yet persistently predicts an attacker-defined label for any sample in the presence of the backdoor trigger. Although backdoor attacks have been extensively studied in the image domain, there are very few works that explore such attacks in the video domain, and they tend to conclude that image backdoor attacks are less effective in the video domain. In this work, we revisit the traditional backdoor threat model and incorporate additional video-related aspects to that model. We show that poisoned-label image backdoor attacks could be extended temporally in two ways, statically and dynamically, leading to highly effective attacks in the video domain. In addition, we explore natural video backdoors to highlight the seriousness of this vulnerability in the video domain. And, for the first time, we study multi-modal (audiovisual) backdoor attacks against video action recognition models, where we show that attacking a single modality is enough for achieving a high attack success rate.
\end{abstract}

\section{Introduction}
\label{sec:introduction}

A fundamental requirement for the deployment of deep neural networks (DNNs) in real-world tasks is their safety and robustness against possible vulnerabilities and security breaches.
This requirement is, in essence, the motivation behind exploring adversarial attacks. One particularly interesting adversarial attack is ``backdoor attacks".
Backdoor attacks or neural trojan attacks explore the scenario in which a user with limited computational capabilities downloads pretrained DNNs from an untrusted party or outsources the training procedure to such a party that we refer to as the adversary. The adversary provides the user with a model that performs well on an unseen validation set, but produces a pre-defined class label in the presence of an attacker-defined  trigger called the backdoor trigger. The association between the backdoor trigger and the attacker-specified label is created by training the DNN on poisoned training samples, which are samples polluted by the attacker's trigger \cite{Li2022BackdoorLA}. In poisoned-label attacks, unlike clean-label attacks, the attacker also switches the label of the poisoned samples to the intended target label.

Considerable attention has been paid to explore backdoor attacks and defenses for 2D image classification models \cite{Barni2019ANB,gu2019badnets,Hammoud2021CheckYO}. However, little attention has been paid to exploring backdoor attacks and defenses against video action recognition models. The disappointing conclusion uncovered by \cite{videoattack} regarding the limited effectiveness of image backdoor attacks on videos stunted further development of video backdoor attacks. Unfortunately, the attacks considered in \cite{videoattack} were limited to only visible patch-based clean-label attacks. Moreover, \cite{videoattack} directly adopted the 2D backdoor attack threat model without incorporating important video-specific considerations.  


To this end, and as opposed to \cite{videoattack}. we first revisit and revise the commonly adopted 2D \emph{poisoned-label} backdoor threat model by incorporating additional constraints that are inherently imposed by video systems. These constraints arise due to the presence of the temporal dimension. We then explore two ways to extend image backdoor attacks to incorporate the temporal dimension into the attack to enable more video-specific backdoor attacks. In particular, image backdoor attacks could be either extended statically by applying the same attack to each frame of the video or dynamically by adjusting the attack parameters differently for each frame. Then, three novel natural video backdoor attacks are presented to highlight the seriousness of the risks associated with backdoor attacks in the video domain. We then test the attacked models against three 2D backdoor defenses and discuss the reason behind the failure of those methods. We also study, for the first time, audiovisual backdoor attacks, where we ablate the importance and contribution of each modality on the performance of the attack for both late and early fusion settings. We show that attacking a single modality is enough to achieve a high attack success rate.

\noindent\textbf{Contributions.} Our contributions are twofold. \textbf{(1)} We revisit the traditional backdoor attack threat model and incorporate video-related aspects, such as video subsampling and spatial cropping, into the model. We also extend existing image backdoor attacks to the video domain in two different ways, statically and dynamically, after which we propose three novel natural video backdoor attacks. Through extensive experiments, we provide evidence that the previous perception of image backdoor attacks in the video domain is not necessarily true, especially in the poisoned-label attack setup. \textbf{(2)} To the best of our knowledge, this work is the first to investigate audiovisual backdoor attacks against video action recognition models.

\begin{figure*}[h!]
    \centering
    \includegraphics[width=0.95\textwidth]{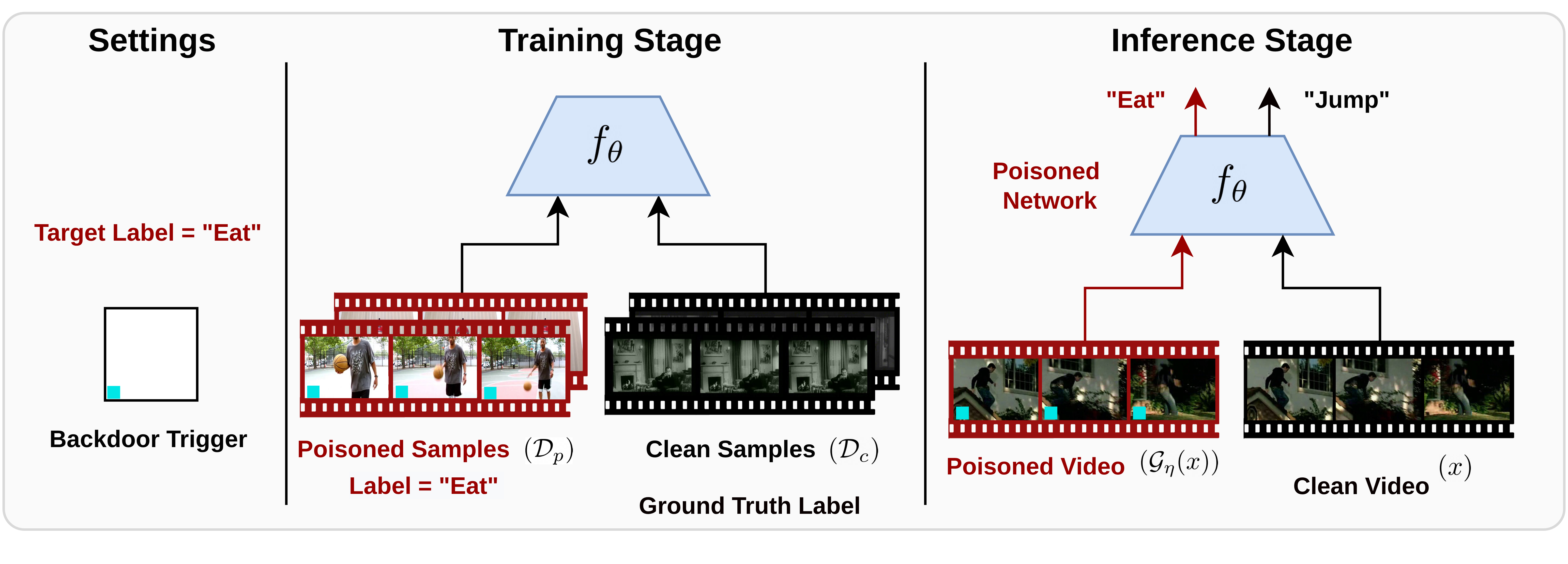}
    \caption{\textbf{Traditional Backdoor Attack Pipeline.} After selecting a backdoor trigger and a target label, the attacker poisons a subset of the training data referred to as the poisoned dataset ($\mathcal{D}_p$). The label of the poisoned dataset is fixed to a target poisoning label specified by the attacker. The attacker trains jointly on clean (non-poisoned) samples ($\mathcal{D}_c$) and poisoned samples leading to a backdoored model, which outputs the target label in the presence of the backdoor trigger.}
    \label{pipeline}
\end{figure*}

\section{Related Work}
\label{sec:relatedwork}

\noindent\textbf{Backdoor Attacks.}
Backdoor attacks were first introduced in \cite{gu2019badnets}. The attack, called BadNet, was based on adding a patch to the corner of a subset of training images to create a backdoor that could be triggered by the attacker at will. Following BadNet, \cite{liu2017trojaning} proposed optimizing for the values of the patch to obtain a more effective backdoor attack. Shortly after the development of patch-based backdoor attacks, the community realized the importance of adding an invisibility constraint to the design of backdoor triggers to bypass any human inspection. Works such as \cite{chen2017targeted} proposed blending the backdoor trigger with the image rather than stamping it. \cite{Li2021InvisibleBA} generated backdoor attacks using the least significant bit algorithm.
\cite{Nguyen2021WaNetI} generated warping fields to warp the image content as a backdoor trigger. \cite{Doan2021LIRALI} went one step further and designed learnable transformations to generate optimal backdoor triggers. After many attacks were proposed in the spatial domain \cite{Liao2020BackdoorEI,Ren2021SimtrojanSB,Chen2021UsePN,Liu2020ReflectionBA,Turner2019LabelConsistentBA,Li2021InvisibleBA,Salem2022DynamicBA,Wang2022BppAttackSA,Xia2022EnhancingBA}, and others in the latent representation domain \cite{Yao2019LatentBA,Qi2022CircumventingBD,Doan2021BackdoorAW,Zhong2022ImperceptibleBA,Zhao2022DEFEATDH}, \cite{Hammoud2021CheckYO,Zeng2021RethinkingTB,Feng2022FIBAFB,Wang2021BackdoorAT,Yue2022InvisibleBA} proposed to switch attention to the frequency domain. \cite{Hammoud2021CheckYO} utilized frequency heatmaps proposed in \cite{Yin2019AFP} to create backdoor attacks that target the most sensitive frequency components of the network. \cite{Feng2022FIBAFB} proposed blending low frequency content from a trigger image with training images as a poisoning technique. \textit{In our work, we extend the 2D backdoor threat model to the video domain by incorporating video-related aspects into it. We also extend five image backdoor attacks into the video domain and propose three natural video backdoor attacks.}

\noindent\textbf{Backdoor Defenses.}
Backdoor attack literature was immediately opposed by various defenses. Backdoor defenses are generally of five types: preprocessing-based \cite{doan2020februus,liu2017neural,qiu2021deepsweep}, model reconstruction-based \cite{liu2018fine,zheng2022data,wu2021adversarial,li2021neural,zeng2021adversarial}, trigger synthesis-based \cite{guo2020towards,shen2021backdoor,hu2021trigger,tao2022better,guo2019tabor,qiao2019defending,wang2019neural,liu2019abs}, model diagonsis-based \cite{xiang2022post,liu2022complex,dong2021black,kolouri2020universal,zheng2021topological}, and sample-filtering based \cite{chen2018detecting,tang2021demon,javaheripi2020cleann,tran2018spectral,gao2019strip,hayase2021spectre}. Early backdoor defenses such as \cite{wang2019neural} hypothesized that backdoor attacks create a shortcut between all samples and the poisoned class. Based on that, they solved an optimization problem to find whether a trigger of an abnormally small norm exists that would flip all samples to one label. Later, multiple improved iterations of this method were proposed, such as \cite{liu2019abs,guo2019tabor,zeng2021adversarial}. Fine pruning \cite{liu2018fine} suggested that the backdoor is triggered by particular neurons that are dormant in the absence of the trigger. Therefore, the authors proposed pruning the least active neurons on clean samples. STRIP \cite{gao2019strip} showed that blending clean samples with other clean samples would yield a higher entropy compared to when clean images are blended with poisoned samples. Activation clustering \cite{chen2018detecting} uses KMeans to cluster the activations of an inspection, a potentially poisoned data set, into two clusters. A large silhouette distance between the two clusters would uncover the poisoned samples. \textit{In our work, we show that current image backdoor attacks have limited effectiveness in defending against backdoor attacks in the video domain, especially against the proposed natural video attacks.}

\noindent\textbf{Video Action Recognition.}
Video action recognition models, which only leverage the raw frames of a video, can be categorized into two categories, CNN-based networks and transformer-based networks. 2D CNN-based methods are built on top of pretrained image recognition networks with well-designed modules to capture the temporal relationship between multiple frames \cite{wang2016temporal, lin2019tsm, luo2019grouped, wang2021tdn}. Those methods are computationally efficient as they use 2D convolutional kernels. To learn stronger spatial-temporal representations, 3D CNN-based methods were proposed. These methods utilize 3D kernels to jointly leverage the spatio-temporal context within a video clip \cite{tran2015learning, feichtenhofer2019slowfast,Feichtenhofer2020X3DEA,Tran2019VideoCW}. To better initialize the network, I3D \cite{carreira2017quo} inflated the weights of 2D pretrained image recognition models to adapt them to 3D CNNs. Realizing the importance of computational efficiency, S3D \cite{xie2018rethinking} and R(2+1)D \cite{tran2018closer} proposed to disentangle spatial and temporal convolutions to reduce computational cost. Recently, transformer-based action recognition models were able to achieve better performance in large training data sets compared to CNN-based models, \eg \cite{arnab2021vivit, fan2021multiscale, liu2022video,Bertasius2021IsSA}. \textit{In this work, we test backdoor attacks against three action recognition architectures, namely I3D, SlowFast, and TSM. }

\noindent\textbf{Audiovisual Action Recognition.}
In addition to frames, a line of action recognition models \cite{Hu2019DeepMC,Morgado2021RobustAI,Hu2020CurriculumAL,Alwassel2020SelfSupervisedLB} has used the accompanying audio to better understand activities such as ``playing music" or ``washing dishes".
To take advantage of existing CNN and transformer-based models, the Log-Mel spectrogram was introduced to convert audio data from a non-structured signal into a 2D representation in time and frequency usable by these models \cite{arandjelovic2017look,arandjelovic2018objects,korbar2018cooperative,xiao2020audiovisual}. Current audiovisual action recognition methods are divided into two categories based on when the audio and visual signals are merged in the recognition pipeline: early fusion and late fusion. Early fusion combines features before classification, which can better capture features \cite{kazakos2019epic, xiao2020audiovisual}. The disadvantage of early fusion is that there is a higher risk of overfitting to the training data \cite{Song2019ARO}. Late fusion, on the other hand, treats the video and audio networks separately, and the predictions of each network are carried out independently, after which the logits are aggregated to make a final prediction \cite{ghanem2018activitynet}. \textit{For the first time, we test backdoor attacks against audiovisual action recognition networks in both late and early fusion setups.}


\begin{figure*}[!htbp]
    \centering
    \includegraphics[width=1.0\textwidth]{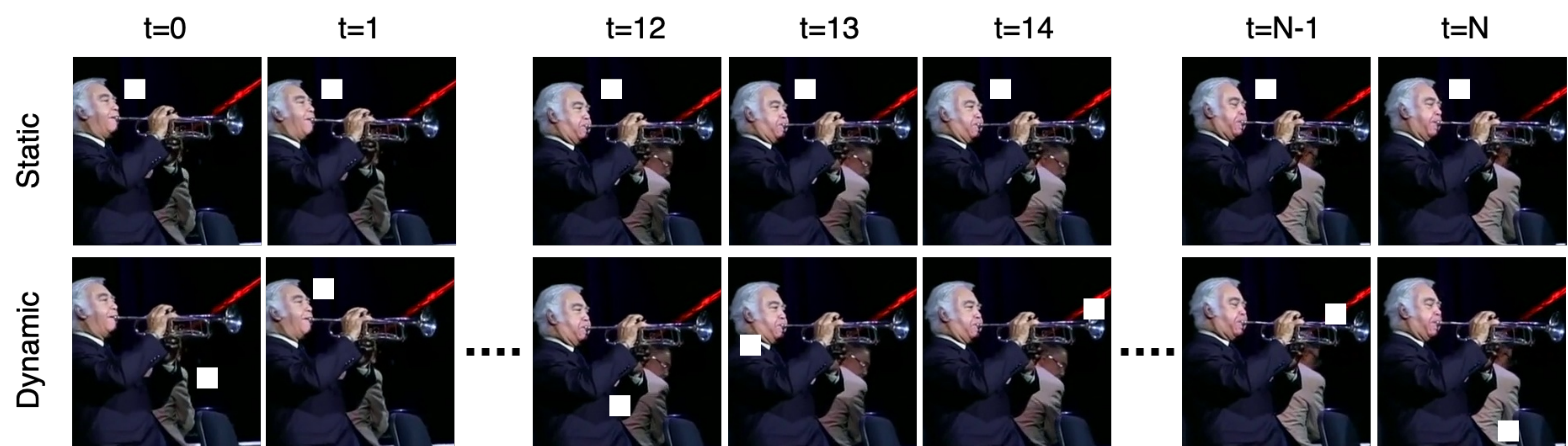}
    \caption{\textbf{Static vs Dynamic Backdoor Attacks.} Static backdoor attacks apply the same trigger across all frames along the temporal dimension. On the other hand, dynamic attacks apply a different trigger per frame along the temporal dimension.}
    \label{fig:dynamicvsstatic}
\end{figure*}

\section{Video Backdoor Attacks}
\label{sec:methodology}

\subsection{The Traditional Threat Model}
\label{formulation}
The commonly adopted threat model for backdoor attacks dates back to the works that studied those attacks against 2D image classification models \cite{gu2019badnets}. The victim outsources the training process to a trainer who is given access to both the victim's training data and the network architecture. The victim only accepts the model provided by the trainer if it performs well on the victim's private validation set. The attacker aims to maximize the effectiveness of the embedded backdoor attack \cite{Li2022BackdoorLA}. We refer to the model's performance on the validation set as clean data accuracy (CDA). The effectiveness of the backdoor attack is measured by the attack success rate (ASR), which is defined as the percentage of test examples not labeled as the target class that are classified as the target class when the backdoor pattern is applied. To achieve this goal, the attacker applies a backdoor trigger to a subset of the training images and then, in the poisoned-label setup, switches the labels of those images to a target class of choice before training begins. A more powerful backdoor attack is one that is visually imperceptible (usually measured in terms of $\ell_2/\ell_\infty$-norm, PSNR, SSIM, or LPIPS) but achieves both a high CDA and a high ASR. This is summarized in Figure \ref{pipeline}.

More formally, we denote the classifier which is parameterized by $\theta$ as $f_\theta: \mathcal{X}\rightarrow \mathcal{Y}$. It maps the input $x\in \mathcal{X}$, such as images or videos, to class labels $y \in \mathcal{Y}$. Let $\mathcal{G}_\eta : \mathcal{X} \rightarrow \mathcal{X}$ indicate an attacker-specific poisoned image generator that is parameterized by some trigger-specific parameters $\eta$. The generator may be image-dependent. Finally, let $\mathcal{S}:\mathcal{Y}\rightarrow \mathcal{Y}$ be an attacker-specified label shifting function. In our case, we consider the scenario in which the attacker is trying to flip all the labels into one particular label, \ie $\mathcal{S}:\mathcal{Y} \rightarrow t $, where $t \in \mathcal{Y}$ is an attacker-specified label that will be activated in the presence of the backdoor trigger. Let $\mathcal{D} = \left\{(\mathbf{x}_i,y_i)\right\}_{i=1}^{N}$ indicate the training dataset. The attacker splits $\mathcal{D}$ into two subsets, a clean subset $\mathcal{D}_c$ and a poisoned subset $\mathcal{D}_p$, whose images are poisoned by $\mathcal{G}_\eta$ and labels are poisoned by $\mathcal{S}$. The poisoning rate is the ratio $\alpha = \frac{\mid\mathcal{D}_p\mid}{\mid\mathcal{D}\mid}$, generally a lower poisoning rate is associated with a higher clean data accuracy. The attacker typically trains the network by minimizing the cross-entropy loss on $\mathcal{D}_c \cup \mathcal{D}_p$, \ie minimizes $\mathbb{E}_{(\mathbf{x},y)\sim\mathcal{D}_c \cup \mathcal{D}_p} [ \mathcal{L}_{CE}(f_\theta(\mathbf{x}),y)]$. The attacker aims to achieve high accuracy on the user's validation set $\mathcal{D}_{val}$ while being able to trigger the poisoned-label, $t$, in the presence of the backdoor trigger, \ie $f_\theta(\mathcal{G}_\eta(\mathbf{x})) = t ,~\forall x \in \mathcal{X}$ (ideally).

\subsection{From Images to Videos}
\label{sec:imtovid}
Unlike images, videos have an additional dimension, the temporal dimension. This dimension introduces new rules to the game between the attacker and the victim. More precisely, the attacker now has an additional dimension to hide the backdoor trigger, leading to a higher level of imperceptibility. The backdoor attack could be applied to all the frames or a subset of the frames statically, \ie the same trigger is applied to each frame, or dynamically, \ie a different trigger is applied to each frame. On the other hand, the testing pipeline now imposes harsher conditions against the backdoor attack. Video recognition models tend to test the model on multiple sub-sampled clips with various crops \cite{lin2019tsm,carreira2017quo,feichtenhofer2019slowfast} which might, in turn, destroy the backdoor trigger. For example, if the trigger is applied to a single frame, it might not be sampled, and if the trigger is applied to the corner of the image, it might be cropped out. The threat model presented in Subsection \ref{formulation} was directly adopted in \cite{videoattack}, which to the best of our knowledge, is the only previous work that considered backdoor attacks for video action recognition. 

Our work sheds light on the aforementioned video-related aspects. In Section \ref{sec:videoattacks}, we show the effect of the number of frames poisoned on CDA and ASR. We also show how existing 2D methods could be extended both statically and dynamically to suit the video domain. For example, BadNet \cite{gu2019badnets} applies a fixed patch as a backdoor trigger. The patch could be applied \underline{statically} using the same pixel values and the same position along the temporal dimension or applied \underline{dynamically} by changing the position and possibly the pixel values of the patch for each frame. Figure \ref{fig:dynamicvsstatic} shows a BadNet attack when applied in a static and dynamic way. Additionally, we show how simple yet natural video ``artifacts" could be used as backdoor triggers. More specifically, lag in a video, motion blur, and compression glitches could all be used as naturally occurring backdoor triggers.

\subsection{Audiovisual Backdoor Attacks}
\label{sec:audioquestions}
Videos are naturally accompanied by audio signals. Similarly to how the video modality could be attacked, the audio signal could also be attacked. The interesting question that arises is how backdoor attacks would perform in a multi-modal setup. In the experiments of Section \ref{sec:audiovisual}, we answer the following questions: (1) What is the effect of having two attacked modalities on CDA and ASR?; (2) What happens if only one modality is attacked and the other is left clean?; (3) What is the difference in performance between late and early fusion in terms of CDA and ASR? 
\begin{figure*}[t!]
    \centering
    \includegraphics[width=0.9\textwidth]{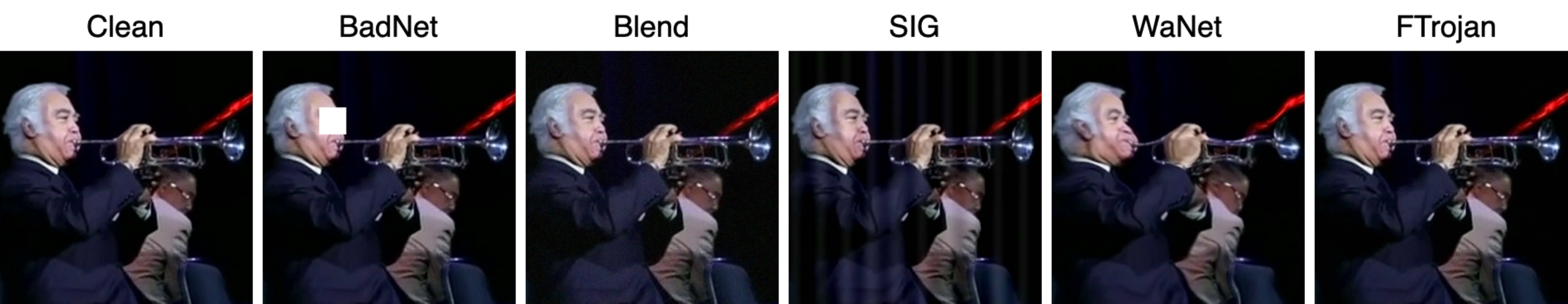}
    \caption{\textbf{Visualization of 2D Backdoor Attacks.} Image backdoor attacks mainly differ according to the backdoor trigger used to poison the training samples. They could  be extended either statically or dynamically based on how the attack is applied across the frames.}
    \label{fig:attacktypes}
\end{figure*}
\section{Experiments}
\label{sec:experiments}

\subsection{Experimental Settings}

\noindent\textbf{Datasets.} We consider three standard benchmark datasets used in video action recognition: UCF-101 \cite{soomro2012ucf101}, HMDB-51 \cite{kuehne2011hmdb}, and Kinetics-Sounds \cite{kay2017kinetics}. Kinetics-Sounds is a subset of Kinetics400 that contains classes that can be classified from the audio signal, \ie classes where audio is useful for action recognition \cite{arandjelovic2017look}. Kinetics-Sounds is particularly interesting for Sections \ref{sec:audio} and \ref{sec:audiovisual}, where we explore backdoor attacks against audio and audiovisual classifiers.

\noindent\textbf{Network Architectures.} Following common practice, for the visual modality, we use a dense sampling strategy to sub-sample 32 frames per video to fine-tune a pretrained I3D network on the target dataset \cite{carreira2017quo}. In Section \ref{sec:videoattacks}, we also show results using TSM \cite{lin2019tsm} and SlowFast \cite{feichtenhofer2019slowfast} networks. All three models adopt ResNet-50 as the backbone and are pretrained on Kinetics-400. Similarly to \cite{arandjelovic2017look}, for the audio modality, a ResNet-18 is trained from scratch on Mel-Spectrograms composed of 80 Mel bands sub-sampled temporally to a fixed length of 256.

\noindent\textbf{Attack Setting.} For the video modality, we study and extend the following image-based backdoor attacks to the video domain: BadNet \cite{gu2019badnets}, Blend \cite{chen2017targeted}, SIG \cite{Barni2019ANB}, WaNet \cite{Nguyen2021WaNetI}, and FTrojan \cite{Wang2021BackdoorAT}. We also explore three additional natural video backdoor attacks. For the audio modality, we consider two attacks: sine attack and high-frequency noise attack, both of which we explain later. Following \cite{gu2019badnets,Hammoud2021CheckYO,Nguyen2021WaNetI}, the target class is arbitrarily set to the first class of each data set (class 0), and the poisoning rate is set to 10\%. Unless otherwise stated, the considered image backdoor attacks poison all frames of the sampled clips during training and evaluation.

\noindent\textbf{Evaluation Metrics.} As is commonly done in the backdoor literature, we evaluate the performance of the model using clean data accuracy (CDA) and attack success rate (ASR) explained in Section \ref{sec:methodology}. CDA represents the usual validation/test accuracy on an unseen dataset hence measuring the generalizability of the model. On the other hand, ASR measures the effectiveness of the attack when the poison is applied to the validation/test set. In addition, we test the attacked models against some of the early 2D backdoor defenses, more precisely against activation clustering (AC) \cite{chen2018detecting}, STRIP \cite{gao2019strip}, and pruning \cite{liu2018fine}.

\noindent\textbf{Implementation Details.} Our method is built on MMAction2 library \cite{2020mmaction2}, and follows their default training configurations and testing protocols, except for the learning rate and the number of training epochs (check Supplementary). All experiments were run using 4 NVIDIA A100 GPUs.

\subsection{Video Backdoor Attacks}
\label{sec:videoattacks}

\noindent\textbf{Extending Image Backdoor Attacks to the Video Domain.}
As mentioned in Section \ref{sec:imtovid}, image backdoor attacks could be extended either statically by applying an attack in the same way across all frames or dynamically by adjusting the attack parameters for different frames. We consider five attacks that differ according to the applied backdoor trigger. BadNet applies a patch as a trigger, Blend blends a trigger image to the original image, SIG superimposes a sinusoidal trigger to the image, WaNet warps the content of the image, and FTrojan poisons a high- and mid- frequency component in the discrete cosine transform (DCT). Figure \ref{fig:attacktypes} visualizes all five attacks on the same video frame. Each of the considered methods could be extended dynamically as follows: \textbf{BadNet}: change the patch location for each frame; \textbf{Blend}: blend a uniform noise that is different per frame; \textbf{SIG}: change the frequency of the sine component superimposed with each frame; \textbf{WaNet}: generate a different warping field for each frame; \textbf{FTrojan}: select a different DCT basis to perturb at each frame. Note that Blend and FTrojan are generally imperceptible. Visualizations and saliency maps for each attack are found in the Supplementary.

Tables \ref{static} and \ref{dynamic} show the CDA and ASR of the I3D models attacked using various backdoor attacks on UCF-101, HMDB-51, and Kinetics-Sounds. Contrary to the conclusion presented in \cite{videoattack}, we find that backdoor attacks are actually highly effective in the video domain. The CDA of the attacked models is very similar to that of the clean unattacked model (baseline), surpassing it in some cases. Extending attacks dynamically, almost always, improves CDA and ASR compared to extending them statically.

\begin{table}[]
\renewcommand{\arraystretch}{1.1}
\scalebox{0.7}{
\begin{tabular}{lcccccc}
\cmidrule(l){2-7}
\textbf{}         & \multicolumn{2}{c}{\textbf{UCF101}} & \multicolumn{2}{c}{\textbf{HMDB51}} & \multicolumn{2}{c}{\textbf{KineticsSound}} \\ \cmidrule(l){2-7} 
\textbf{}         & \textbf{CDA(\%)} & \textbf{ASR(\%)} & \textbf{CDA(\%)} & \textbf{ASR(\%)} & \textbf{CDA(\%)}     & \textbf{ASR(\%)}    \\ \hhline{-------}
\rowcolor[HTML]{EFEFEF} 
\textbf{Baseline} & 93.95            & -                & 69.59            & -                & 81.41                & -                   \\
\textbf{BadNet}   & 93.95            & 99.63            & 69.35            & 98.89            & 82.97                & 99.09               \\
\rowcolor[HTML]{EFEFEF} 
\textbf{Blend}    & 94.29            & 99.26            & 68.37            & 86.73            & 82.12                & 97.54               \\
\textbf{SIG}      & 93.97            & 99.97            & 68.50            & 99.80            & 82.84                & 99.87               \\
\rowcolor[HTML]{EFEFEF} 
\textbf{WaNet}    & 94.05            & 99.84            & 68.95            & 99.61            & 82.38                & 99.09               \\
\textbf{FTrojan}  & 94.16            & 99.34            & 68.10            & 97.52            & 82.45                & 97.86               \\ \hhline{-------}
\end{tabular}}
\caption{\textbf{Statically Extended 2D Backdoor Attacks.} Statically extending 2D backdoor attacks to the video domain leads to high CDA and ASR across all three considered datasets. }
\label{static}
\end{table}

\begin{table}[]
\renewcommand{\arraystretch}{1.1}

\scalebox{0.7}{
\begin{tabular}{lcccccc}
\cmidrule(l){2-7}
\textbf{}         & \multicolumn{2}{c}{\textbf{UCF101}} & \multicolumn{2}{c}{\textbf{HMDB51}} & \multicolumn{2}{c}{\textbf{KineticsSound}} \\ \cmidrule(l){2-7} 
\textbf{}         & \textbf{CDA(\%)} & \textbf{ASR(\%)} & \textbf{CDA(\%)} & \textbf{ASR(\%)} & \textbf{CDA(\%)}     & \textbf{ASR(\%)}    \\ \hhline{-------}
\rowcolor[HTML]{EFEFEF} 
\textbf{Baseline} & 93.95            & -                & 69.59            & -                & 81.41                & -                   \\
\textbf{BadNet}   & 94.11            & 99.97            & 69.08            & 99.54            & 82.25                & 99.74               \\
\rowcolor[HTML]{EFEFEF} 
\textbf{Blend}    & 94.21            & 99.44            & 67.03            & 95.95            & 81.67                & 95.79               \\
\textbf{SIG}      & 94.24            & 100.00           & 68.63            & 100.00           & 82.84                & 100.00              \\
\rowcolor[HTML]{EFEFEF} 
\textbf{WaNet}    & 94.29            & 99.79            & 69.22            & 99.80            & 82.25                & 99.61               \\
\textbf{FTrojan}  & 94.16            & 99.34            & 67.19            & 98.69            & 82.25                & 95.27               \\ \hhline{-------}
\end{tabular}}
\caption{\textbf{Dynamically Extended 2D Backdoor Attacks.} Dynamically extending 2D backdoor attacks to the video domain leads to high CDA and ASR across all three considered datasets. }
\label{dynamic}
\end{table}

\noindent \textbf{Natural Video Backdoors.} A more interesting attack is one that seems natural and could bypass human inspection \cite{Ma2022MACABMC,Xue2021RobustBA,Wenger2022NaturalBD,Zhao2022NaturalBA}. There are several natural ``glitches" that occur in the video domain and that one could exploit to design a natural backdoor attack. For example, videos might contain some frame lag, motion blur, video compression corruptions, camera focus/defocus, etc. In Table \ref{naturalattack}, we report the CDA and ASR of three natural backdoor attacks: frame lag (lagging video), video compression glitch (which we refer to as Video Corruption), and motion blur. Interestingly, these attacks could achieve both high clean data accuracy and high attack success rate. It is worth noting that for frame lag, a two-frame lag is used for UCF-101 and a three-frame lag is used for HMDB-51 and Kinetics-Sounds. More details are provided in the Supplementary.

\begin{table}[t!]
\renewcommand{\arraystretch}{1.2}

\scalebox{0.67}{
\begin{tabular}{lcccccc}
\cmidrule(l){2-7}
\textbf{}             & \multicolumn{2}{c}{\textbf{UCF101}} & \multicolumn{2}{c}{\textbf{HMDB51}} & \multicolumn{2}{c}{\textbf{KineticsSound}} \\ \cmidrule(l){2-7} 
\textbf{}             & \textbf{CDA(\%)} & \textbf{ASR(\%)} & \textbf{CDA(\%)} & \textbf{ASR(\%)} & \textbf{CDA(\%)}     & \textbf{ASR(\%)}    \\ \hhline{-------}
\rowcolor[HTML]{EFEFEF} 
\textbf{Baseline}     & 93.95            & -                & 69.59            & -                & 81.41                & -                   \\
\textbf{Frame Lag}    & 92.94            & 97.20            & 68.04            & 98.76            & 82.51                & 98.19               \\
\rowcolor[HTML]{EFEFEF} 
\textbf{\small Video Corrupt.} & 94.26            & 99.87            & 69.22            & 99.22            & 81.74                & 98.51               \\
\textbf{Motion Blur}   & 93.97            & 99.92            & 68.17            & 97.52            & 82.19                & 99.22   \\   \hhline{-------}        
\end{tabular}}
\caption{\textbf{Natural Video Backdoor Attacks.} Natural attacks against video action recognition models could achieve high CDA and ASR while looking completely natural to human inspection.}
\label{naturalattack}
\end{table}

\noindent\textbf{Attacks Against Different Architectures.} So far, all attacks have been experimented with against an I3D network. To further explore the behavior of backdoor attacks against other video recognition models, we test a subset of the considered attacks against a 2D based model, TSM, and another 3D based model, SlowFast, on UCF-101. Table \ref{arch} shows that all the aforementioned backdoor attacks perform significantly well in terms of CDA and ASR against both TSM and SlowFast architectures. Note that even though TSM is a 2D based model, our proposed natural video backdoor attacks still succeed in attacking it.

\begin{table}[]
\renewcommand{\arraystretch}{1.0}
\centering
\scalebox{0.8}{
\begin{tabular}{lcccc}
\cmidrule(l){2-5}
\multicolumn{1}{c}{}                                                     & \multicolumn{2}{c}{\textbf{SlowFast}} & \multicolumn{2}{c}{\textbf{TSM}}    \\ \cmidrule(l){2-5} 
\multicolumn{1}{c}{}                                                     & \textbf{CDA(\%)}  & \textbf{ASR(\%)}  & \textbf{CDA(\%)} & \textbf{ASR(\%)} \\ \hhline{-----}
\rowcolor[HTML]{EFEFEF} 
\textbf{Baseline}                                                        & 96.72             & -                 & 94.77            & -                \\
\textbf{BadNet}                                                          & 96.64             & 99.47             & 94.69            & 97.78            \\
\rowcolor[HTML]{EFEFEF} 
\textbf{SIG}                                                             & 96.70             & 99.97             & 94.77            & 99.47            \\
\textbf{FTrojan}                                                         & 96.25             & 98.52             & 94.21            & 100.00           \\
\rowcolor[HTML]{EFEFEF} 
\textbf{Frame Lag}                                                       & 96.43             & 99.97             & 94.63            & 97.96            \\
\textbf{\begin{tabular}[c]{@{}l@{}}Video  Corruption\end{tabular}} & 96.54             & 99.76             & 95.08            & 98.97            \\
\rowcolor[HTML]{EFEFEF} 
\textbf{Motion Blur}                                                     & 96.46             & 99.55             & 94.50            & 99.39            \\ \hhline{-----}
\end{tabular}}
\caption{\textbf{Video Backdoor Attacks Against Different Architectures (UCF-101).} When tested against network architectures other than I3D such as TSM and SlowFast, both image and natural backdoor attacks can still achieve high CDA and high ASR.}
\label{arch}
\end{table}

\begin{figure}[t!]
    \centering
    \includegraphics[width=\linewidth]{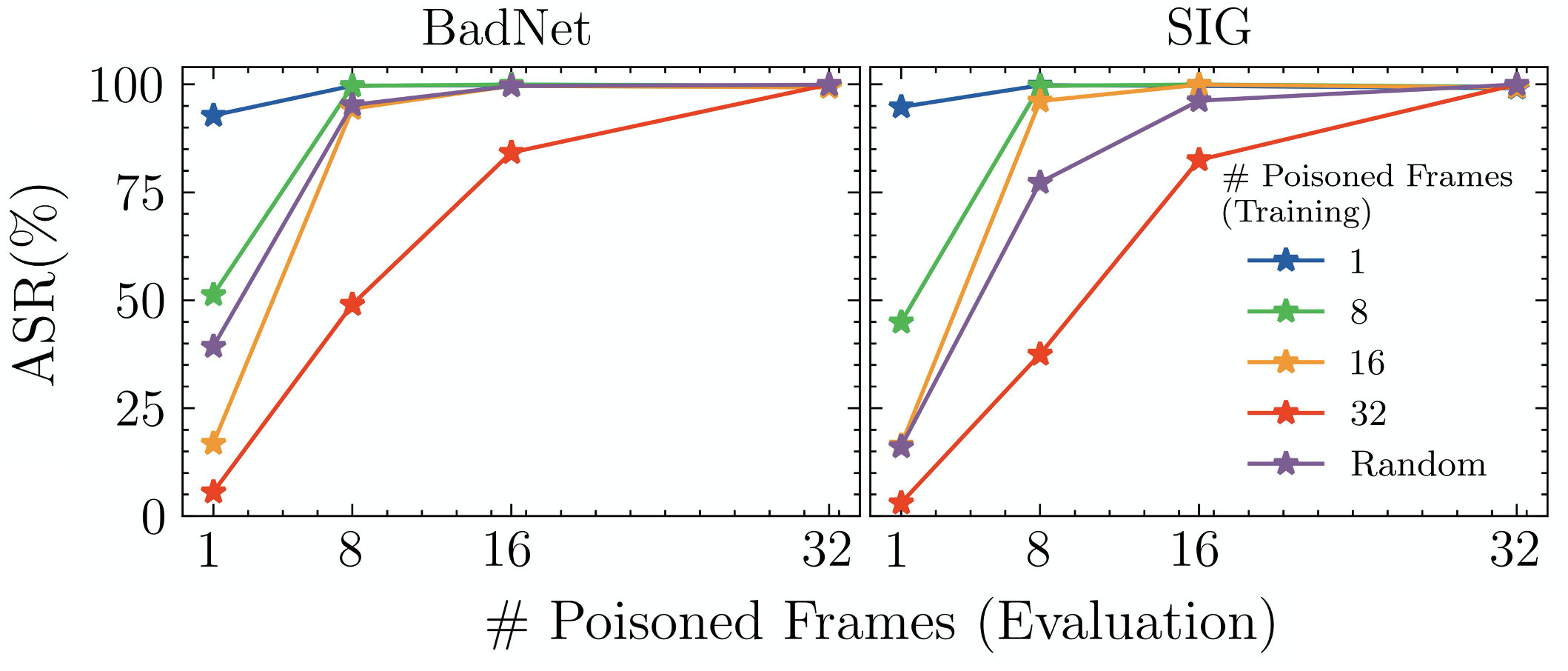}
    \caption{\textbf{Effect of the Number of Poisoned Frames (UCF-101).} Different colors refer to different number of frames poisoned during the training of the attacked model. Training the model with a single poisoned frame performs best for various choices of the number of frames poisoned during evaluation.}
    \label{ablate2}
\end{figure}

\noindent\textbf{Recommendations for Video Backdoor Attacks.}
As mentioned in Section \ref{sec:imtovid}, the attacker must select a number of frames to poison per video, keeping in mind that the video will be sub-sampled and randomly cropped during evaluation. Since the attacker is the one who trained the network in the first place, he/she has access to the processing pipeline and could exploit this during the attack. For example, if video processing involves sub-sampling the video into clips of 32 frames and cropping the frames into 224$\times$224 crops, the attacker could pass to the network an attacked video of a temporal length of 32 frames and a spatial size 224$\times$224, hence bypassing sub-sampling and cropping. However, a system could force the user to input a video of a particular length, possibly greater than the length of the sub-sampled clips. This raises an important question regarding how many frames the attacker should poison. Clearly, the smaller the number of frames the attacker poisons, the less detectable the attack is, but does the attack remain effective? In Figure \ref{ablate2}, we show the attack success rate of backdoor-attacked models \textbf{trained} on clips of 1, 8, 16, and 32 frames, and a randomly sampled number of poisoned frames (out of 32 total frames) when \textbf{evaluated} on clips of 1, 8, 16, and 32 poisoned frames (out of 32 total frames). Random refers to training on a varying number of poisoned frames per clip. Note that training the model against the worst-case scenario (single frame), which mimics the case where only one of the poisoned frames is sub-sampled, provides the best guarantees for achieving a high attack success rate.

\begin{table}[t!]
\scalebox{0.7}{
\begin{tabular}{lccccc}
\cmidrule(l){2-6}
                                                                         & \textbf{\begin{tabular}[c]{@{}c@{}}Frame\\ Lag\end{tabular}} & \textbf{\begin{tabular}[c]{@{}c@{}}Motion\\ Blur\end{tabular}} & \textbf{SIG} & \textbf{BadNet} & \textbf{FTrojan} \\ \midrule
\textbf{\begin{tabular}{l}Elimination Rate(\%)\end{tabular}} & 0.00                                                         & 0.00                                                          & 34.21        & 33.77           & 34.12            \\
\textbf{\begin{tabular}{l}Sacrifice Rate(\%)\end{tabular}}   & 13.08                                                        & 12.82                                                         & 15.17        & 14.25           & 13.00            \\ \bottomrule
\end{tabular}}
\caption{\textbf{Activation Clustering Defense (UCF-101).} Whereas Activation Clustering provides partial success in defending against image backdoor attacks, it fails completely against natural attacks.}
\label{ac}
\end{table}

\noindent\textbf{Defenses Against Video Backdoor Attacks.} 
We explore the effect of extending some of the existing 2D backdoor defenses against video backdoor attacks. Optimization-based defenses are extremely costly when extended to the video domain. For example, Neural Cleanse (NC) \cite{wang2019neural}, I-BAU \cite{zeng2021adversarial}, and TABOR \cite{guo2019tabor}  involve a trigger reconstruction phase. The trigger space is now bigger in the presence of the temporal dimension, and therefore, instead of optimizing for a 224$\times$224$\times$3 trigger, the defender has to search for a 32$\times$224$\times$224$\times$3 trigger (assuming 32 frame clips are used), which is both costly and hard to solve. The attacker has the spatial and temporal dimensions to design and embed their attack in, and, therefore, reverse engineering the trigger is quite hard.  

\begin{figure}[t!]
    \centering
    \includegraphics[width=0.8\linewidth]{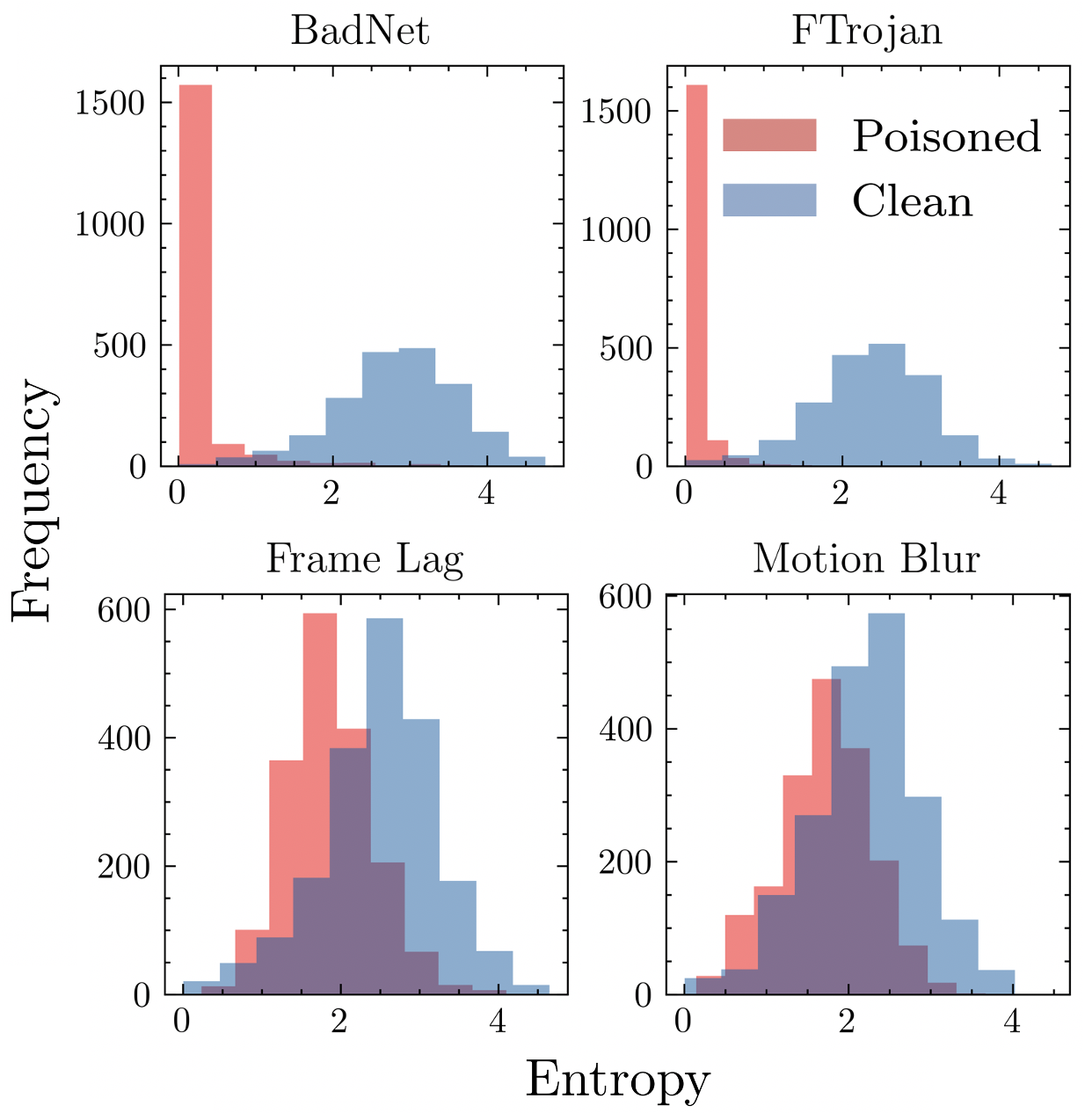}
    \caption{{\textbf{STRIP Defense (UCF-101).}} Whereas the entropy of image backdoor attacks is very low compared to that of clean samples, the proposed natural backdoor attacks have a natural distribution of entropies similar to that of clean samples.}
    \label{strip}
\end{figure}

\begin{figure}[t!]
    \centering
    \includegraphics[width=\linewidth]{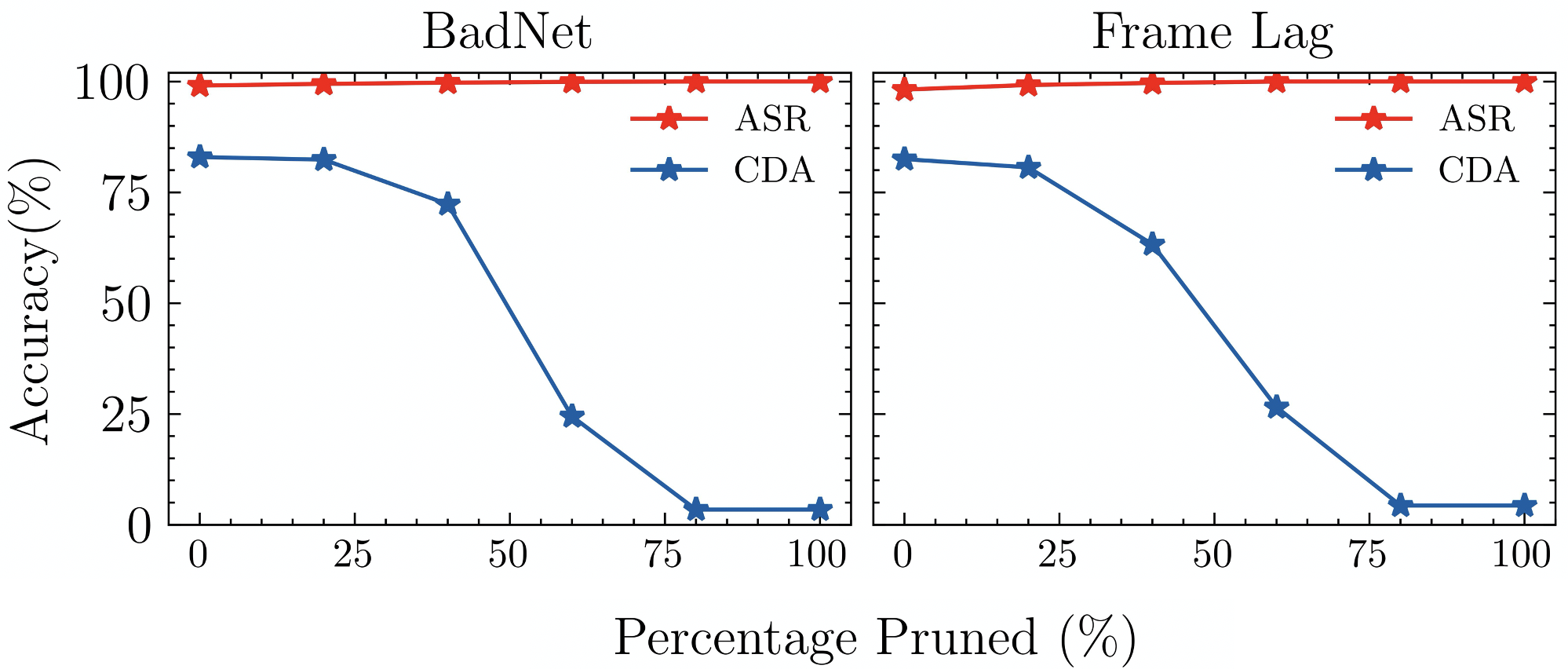}
    \caption{\textbf{Pruning Defense (Kinetics-Sounds).} Pruning is completely ineffective against image backdoor attacks extended to the video domain and natural video backdoor attacks. Even though the clean accuracy has dropped to random, the attack success rate is maintained at very high levels.}
    \label{pruning}
\end{figure}

We consider three well-known defenses that introduce no computational overhead when adopted to the video domain, namely Activation Cluster (AC) \cite{chen2018detecting}, STRIP \cite{gao2019strip}, and pruning \cite{liu2018fine}. 
AC computes the activations of a neural network on clean samples (from the test set) and an inspection set of interest which may be poisoned. AC then applies PCA to reduce the dimension of the activations, after which the projected activations are clustered into two classes and compared to the activations of the clean set. STRIP blends clean samples with the samples of a possibly poisoned inspection set. The entropy of the predicted probabilities is then checked for any abnormalities. Unlike clean samples, poisoned samples tend to have a low entropy. Pruning suggests that the backdoor is usually embedded in particular neurons in the network that are only activated in the presence of the trigger. Therefore, those neurons are supposed to be dormant as far as the test set samples, \ie clean samples, are concerned. This allows us to detect and prune those dormant neurons to eliminate the backdoor. 
Table \ref{ac} shows the elimination and sacrifice rates of AC when applied against some of the considered attacks. The elimination rate refers to the ratio of poisoned samples correctly detected as poisoned to the total number of poisoned samples, whereas the sacrifice rate refers to the ratio of clean samples incorrectly detected as poisoned to the total number of clean samples. Whereas AC has partial success in defending against image backdoor attacks, it fails completely against the proposed natural backdoor attacks. Figure \ref{strip} shows that the entropy of the clean and poisoned samples of the proposed natural attacks is very similar and therefore could evade the STRIP defense, while BadNet and FTrojan are detectable. Finally, Figure \ref{pruning} shows that pruning the least active neurons causes a reduction in CDA without reducing ASR. This is observed not only for the natural attacks, but also for the extended image backdoor attacks, hinting that image backdoor defenses are not effective in the video domain.


\begin{figure}[!t]
    \centering
    \includegraphics[width=0.83\linewidth]{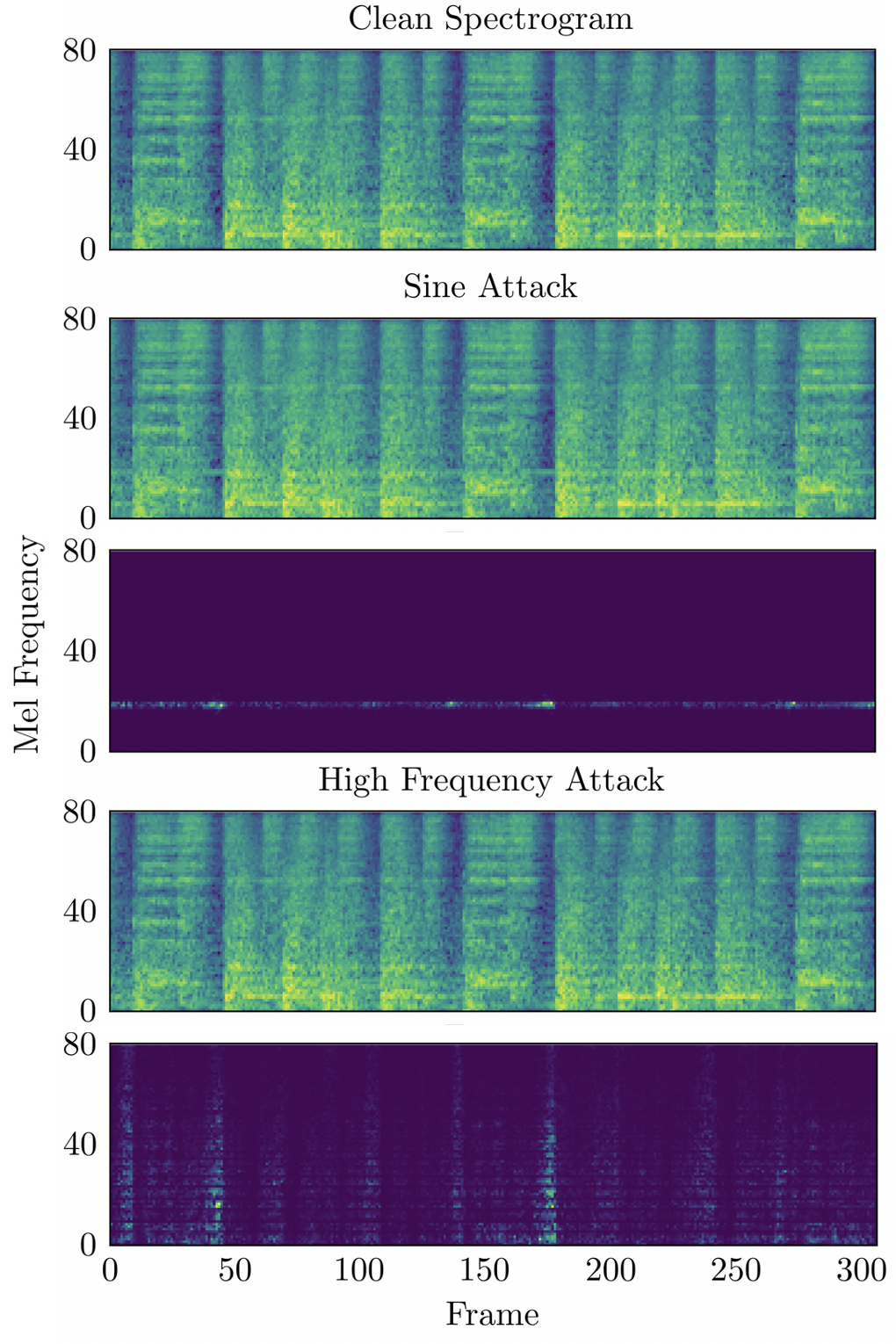}
    \caption{\textbf{Clean and Attacked Audio Spectrograms.} The utilized audio backdoor attacks are not only audibly imperceptible but also leave no perceptible artifacts in the Mel spectrogram. The spectrogram of each attack is followed by the absolute difference of the attacked spectrogram with the clean one.}
    \label{spectrogram}
\end{figure}
\begin{table}[t!]
\centering
\scalebox{0.8}
{
\begin{tabular}{@{}lccc@{}}
\hhline{~---}
                 & \multicolumn{1}{l}{\textbf{Baseline}} & \multicolumn{1}{l}{\textbf{Sine Attack}} & \multicolumn{1}{l}{\textbf{High Frequency Attack}} \\ \midrule
\textbf{CDA(\%)} & 49.21                                 & 47.21                                    & 47.61                                              \\
\textbf{ASR(\%)} & -                                     & 96.36                                    & 95.96                                              \\ \bottomrule
\end{tabular}}
\caption{\textbf{Audio Backdoor Attacks (Kinetics-Sounds).} Both sine attack and the high-frequency band attack perform similarly to baseline in terms of CDA while being able to achieve high ASR.}
\label{audioattack}
\end{table}

\begin{table*}[!t]
\centering
\renewcommand{\arraystretch}{1.2}
\scalebox{0.9}{
    \begin{tabular}{lccc||ccc}
    \cmidrule(l){2-7}
    \multicolumn{1}{c}{}  & \multicolumn{3}{c}{\textbf{Late Fusion}}                                                                               & \multicolumn{3}{c}{\textbf{Early Fusion}}                                                                                \\ \cmidrule(l){2-7} 
    \multicolumn{1}{c}{}  & \textbf{\small Clean Audio} & \textbf{\small Sine Attack} & \textbf{\begin{tabular}[c]{@{}c@{}}{\small High Freq.  Attack}\end{tabular}} & \textbf{\small Clean Audio} & \textbf{\small Sine Attack} & \textbf{\begin{tabular}[c]{@{}c@{}}{\small High Freq.  Attack}\end{tabular}} \\ \hhline{-------}
    \rowcolor[HTML]{EFEFEF} 
    \textbf{Clean Video}  & 80.25 / -              & 81.74 / 70.98          & 80.96 / 77.91                                                               & 84.72 / -              & 83.48 / 92.23          & 83.94 / 93.72                                                               \\
    \textbf{BadNet}       & 77.33 / 66.97          & 78.63 / 99.74          & 77.33 / 99.87                                                               & 87.50 / 99.29          & 85.10 / 99.87          & 85.75 / 100.00                                                              \\
    \rowcolor[HTML]{EFEFEF} 
    \textbf{Blend}        & 79.60 / 75.06          & 80.76 / 99.68          & 79.08 / 99.61                                                               & 86.08 / 98.19          & 83.55 / 99.81          & 85.43 / 99.87                                                               \\
    \textbf{SIG}          & 78.50 / 68.33          & 80.12 / 99.87          & 79.02 / 100.00                                                              & 86.92 / 99.81          & 84.97 / 100.00         & 85.95 / 100.00                                                              \\
    \rowcolor[HTML]{EFEFEF} 
    \textbf{WaNet}        & 77.66 / 68.39          & 79.79 / 99.94          & 79.02 / 99.94                                                               & 86.46 / 98.96          & 84.97 / 100.00         & 85.88 / 100.00                                                              \\
    \textbf{FTrojan}      & 79.66 / 67.16          & 80.76 / 99.48          & 79.99 / 99.29                                                               & 86.08 / 98.58          & 84.65 / 99.94          & 85.49 / 100.00                                                              \\
    \rowcolor[HTML]{EFEFEF} 
    \textbf{Frame Lag}    & 79.08 / 63.41          & 80.57 / 99.74          & 79.47 / 99.87                                                               & 86.08 / 98.19          & 84.59 / 99.94          & 84.65 / 100.00                                                              \\
    \textbf{Video Corruption} & 78.11 / 64.57          & 78.24 / 99.68          & 77.66 / 99.94                                                               & 86.59 / 99.29          & 84.59 / 100.00         & 85.43 / 100.00                                                              \\
    \rowcolor[HTML]{EFEFEF} 
    \textbf{Motion Blur}   & 79.79 / 69.24          & 80.70 / 99.68          & 79.86 / 99.94                                                               & 86.40 / 98.58          & 84.65 / 100.00         & 85.62 / 100.00                                                              \\ \bottomrule
    \end{tabular}}
    \caption{\textbf{Audiovisual Backdoor Attacks (Kinetics-Sounds).} The entries in the table report the CDA(\%)/ASR(\%) of attacking late and early fused audiovisual networks. When a single modality is attacked, late fusion has a low ASR compared to early fusion. When both modalities are attacked, the ASR of both late and early fusion are high.  }
    \label{avresults}
    \end{table*}

\subsection{Audio Backdoor Attacks}
\label{sec:audio}
Attacks proposed against audio networks have been limited to adding a low-volume one-hot-spectrum noise in the frequency domain, which leaves highly visible artifacts in the spectrogram \cite{Zhai2021BackdoorAA} or adding a human non-audible component \cite{Koffas2022CanYH}, $f<20\text{Hz}$ or $f>20\text{kHz}$, which is non-realistic, since spectrograms usually filter out those frequencies. We consider two attacks against the Kinetics-Sounds dataset; the first is to add a low-amplitude sine wave component with $f=800\text{Hz}$ to the audio signal, and the second is to add band-limited noise $5\text{kHz}<f<6\text{kHz}$. The spectrograms and the absolute difference between the attacked spectrograms and the clean spectrogram are shown in Figure \ref{spectrogram}. Since no clear artifacts are observed in the spectrograms, human inspection fails to label the spectrograms as attacked. The CDA and ASR rates of the backdoor-attacked models for both attacks are shown in Table \ref{audioattack}. The attacks achieve a relatively high ASR.

\subsection{Audiovisual Backdoor Attacks}
\label{sec:audiovisual}
Now, we combine video and audio attacks to build a multi-modal audiovisual backdoor attack. The way we do it is by taking our attacked models from Sections \ref{sec:videoattacks} and \ref{sec:audio} and applying early or late fusion. For early fusion, we extract video and audio features using our trained audio and video backbones, and we then train a classifier on the concatenation of the features. In late fusion, the video and audio networks predict independently on the input, and then the individual logits are aggregated to produce the final prediction. To answer the three questions posed in Section \ref{sec:audioquestions}, we run experiments in which both modalities are attacked and others in which only a single modality is attacked for both early and late fusion setups (Table \ref{avresults}). We summarize the results as follows. \textbf{(1)} Attacking two modalities consistently improves ASR and even CDA in some cases. \textbf{(2)} Attacking a single modality is good enough to achieve a high ASR in the case of early fusion but not late fusion. \textbf{(3)} Early fusion enables the best of both worlds for the attacker, namely, a high CDA and an almost perfect ASR. On the other hand, late fusion experiences some serious drops in ASR in the unimodal attack setup. An interesting finding in these experiments is the following: if the outsourcer has the option to outsource the most expensive modality, training wise, while training other modalities in-house, applying late fusion could be used as a defense mechanism, especially in the presence of more clean modalities.

\section{Conclusion}
\label{sec:conclusion}

Backdoor attacks present a serious and exploitable vulnerability against both unimodal and multi-modal video action recognition models. We showed how existing image backdoor attacks could be extended either statically or dynamically to develop powerful backdoor attacks that achieve both a high clean data accuracy and a high attack success rate. Besides existing image backdoor attacks, there exists a set of natural video backdoor attacks, such as motion blur and frame lag, that are resilient to existing image backdoor defenses. Given that videos are usually accompanied by audio, we showed two ways in which one could attack audio classifiers in a human inaudible manner. The attacked video and audio models are then used to train an audiovisual action recognition model by applying both early and late fusion. Different combinations of poisoned modalities are tested, concluding that: \textbf{(1)} poisoning two modalities could achieve extremely high attack success rates in both late and early fusion settings, and \textbf{(2)} if a single modality is poisoned, unlike early fusion, late fusion could reduce the effectiveness of the backdoor. We hope that our work reignites the attention of the community towards exploring backdoor attacks and defenses in the video domain.

{\small
\bibliographystyle{ieee_fullname}
\bibliography{egbib}
}

\end{document}